*Evgenii Evstafev*

# Benchmarking Multimodal Models for Fine-Grained Image Analysis: A Comparative Study Across Diverse Visual Features.

Evgenii Evstafev [A]

[A] University Information Services (UIS), University of Cambridge,
Roger Needham Building, 7 JJ Thomson Ave, Cambridge CB3 0RB, UK, ee345@cam.ac.uk

**ABSTRACT**

*This article introduces a benchmark designed to evaluate the capabilities of multimodal models in analyzing and interpreting images. The benchmark focuses on seven key visual aspects: main object, additional objects, background, detail, dominant colors, style, and viewpoint. A dataset of 14,580 images, generated from diverse text prompts, was used to assess the performance of seven leading multimodal models. These models were evaluated on their ability to accurately identify and describe each visual aspect, providing insights into their strengths and weaknesses for comprehensive image understanding. The findings of this benchmark have significant implications for the development and selection of multimodal models for various image analysis tasks.*

**TYPE OF PAPER AND KEYWORDS**

*Benchmarking, multimodal models, image analysis, computer vision, deep learning, image understanding, visual features, model evaluation*

## 1 INTRODUCTION

Multimodal models, capable of processing and integrating information from multiple modalities such as text and images [1], have emerged as a powerful tool for comprehensive image understanding [2]. These models hold the potential to revolutionize various applications, including image retrieval, content creation, and human-computer interaction. However, evaluating their ability to capture fine-grained details and contextual information remains a crucial challenge [3]. This article presents a benchmark for evaluating the performance of different multimodal models in identifying and analyzing specific aspects of images, such as the main object, additional objects, background, details, dominant colors, style, and viewpoint. By comparing their performance across a range of tasks, this research aims to provide insights into the strengths and weaknesses of different multimodal approaches for fine-grained image analysis.

## 2. BACKGROUND AND RELATED WORK

Multimodal models in computer vision use the interplay between different modalities, such as text and images, to achieve a more holistic understanding of visual content [4]. This approach has shown promising results in various tasks, including image captioning, visual question answering, and image generation [5]. Recent advancements in deep learning, particularly the development of transformer-based architectures, have further accelerated progress in this field. Models like CLIP (Contrastive Language-Image Pre-training [6]) have demonstrated the ability to learn robust representations that capture the semantic relationship between images and text [7]. However, there is a need for standardized benchmarks to evaluate the performance of multimodal models in fine-grained image analysis, as existing benchmarks often focus on broader tasks without explicitly assessing their ability to capture subtle details and contextual information [8]. This research addresses this gap by introducing a benchmark that specifically targets the analysis of diverse visual features in images, enabling a more comprehensive evaluation of their capabilities.

## 3. METHODOLOGY

### 3.1 DATASET CREATION

The dataset creation process involved generating a diverse set of image descriptions (prompts) by systematically combining distinct visual aspects. These aspects, along with the number of unique variations for each, were predefined as follows:

1. **Styles (3 variations)**: Artistic or visual styles of the image (e.g., "Glitch-Infused Surrealism").
2. **Backgrounds (3 variations):** The scene or environment surrounding the objects (e.g., "Desolate Moorland Under a Cracked Sky").
3. **Viewpoints (3 variations):** The perspective from which the image is captured (e.g., "Inside a Mirrored Kaleidoscope").
4. **Main Objects (5 variations):** The primary subject of the image (e.g., "Bioluminescent Coral Spire").
5. **Additional Objects (3 variations):** Secondary objects present in the image (e.g., "Intricate Clockwork Dragonfly").
6. **Details (3 variations):** Specific features or characteristics of the objects or background (e.g., "Pulsating Neon Circuitry").
7. **Colors (3 variations):** The most prominent colors in the image (e.g., "Electric Ultraviolet Haze").

All possible combinations of these aspects were generated, resulting in 3,645 (3 x 3 x 3 x 5 x 3 x 3 x 3) unique text prompts [9]. For each prompt, four images were generated using the flux.1-pro model [10] with different random seeds, resulting in a total of 14,580 images [11]. Each prompt followed a predefined template [12], incorporating all the visual aspects in a descriptive sentence. For example:

*"Glitch-Infused Surrealism style image depicting Desolate Moorland Under a Cracked Sky from an Inside a Mirrored Kaleidoscope. The scene includes Bioluminescent Coral Spire and Intricate Clockwork Dragonfly, focusing on Pulsating Neon Circuitry. The palette is dominated by Electric Ultraviolet Haze."*

### 3.2 MODEL SELECTION

This study employed seven multimodal models with vision capabilities, chosen for their relevance to the research and availability as of January 13, 2025:

- **claude-3-5-sonnet-20241022**: The most recent and accessible model from Anthropic as of the aforementioned date [13].
- **minicpm-v:8b**: A model within the 7b to 11b size range [14], suitable for local execution via Ollama [15].
- **gpt-4o-mini-2024-07-18**: The latest accessible model in the GPT-4 series [16].
- **pixtral-large-2411** & **pixtral-12b-2409**: Models from Mistral, specifically designed for computer vision experiments [17].
- **llama3.2-vision:11b**: A model within the 7b to 11b size range [18], suitable for local execution via Ollama.
- **llava:7b**: A model within the 7b to 11b size range [19], suitable for local execution via Ollama.

These models were selected based on their diverse architectures, capabilities, and availability for comprehensive evaluation. Each model received the same image prompts with explicit instructions to generate descriptions focusing on the key criteria: main object, additional objects, background, details, dominant colors, style, and viewpoint [20]. This ensured a consistent evaluation framework across all models.

### 3.3 EVALUATION METRICS

To evaluate the performance of the multimodal models, a dedicated evaluation model, "mistral-small-2409," [21] was used. This model was tasked with comparing the original image description with the generated description from each model [22]. The use of a separate evaluation model is crucial to avoid biases and ensure a fair comparison between the models being benchmarked. The comparison was performed based on seven specific criteria:

1. **Style:** The artistic or visual style of the image.
2. **Background:** The scene or environment surrounding the objects.
3. **Viewpoint:** The perspective from which the image is captured.
4. **Main Object:** The primary subject of the image.
5. **Additional Objects:** Secondary objects present in the image.
6. **Detail:** Specific features or characteristics of the objects or background.
7. **Dominant Colors:** The most prominent colors in the image.

For each criterion, the evaluation model assigned a score from 0 to 100, representing the degree of match between the original image description and the generated description [22]. A score of 100 indicated a perfect match, while a score of 0 indicated no match at all. For example, if a model accurately identified the "Bioluminescent Coral Spire" as the main object, it would receive a high score for that criterion. However, if it misidentified the object or failed to mention it, the score would be lower. Intermediate scores reflected the degree of difference between the two descriptions, allowing for a nuanced assessment of the models' performance.

This process allowed for a quantitative assessment of each model's ability to accurately capture and describe the various visual aspects of the images. The scores for each criterion were then aggregated to provide an overall performance measure for each model.

## 4. RESULTS

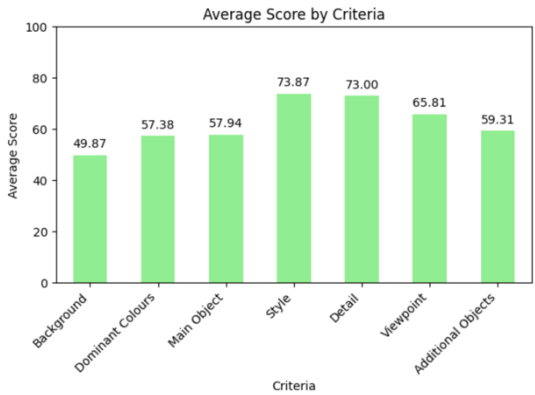

**Chart 1: Model Performance Across Criteria**

- **X-axis:** Criteria (Style, Background, Viewpoint, Main Object, Additional Objects, Detail, Dominant Colors)
- **Y-axis:** Average score for each criterion across all models

Bar chart showing the average performance of all models for each criterion. It highlights which visual aspects are generally well-captured and which pose challenges for the models.

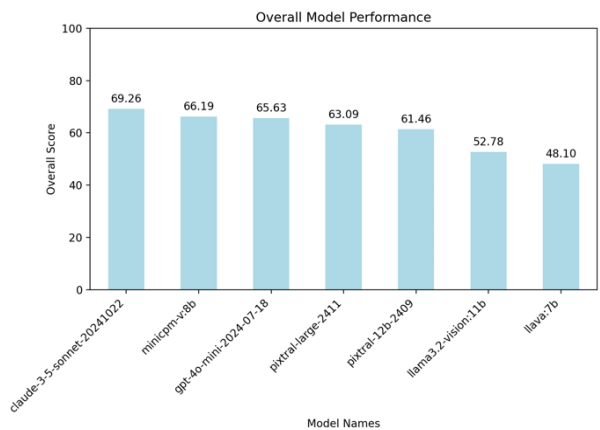

**Chart 2: Overall Model Performance**

- **X-axis:** Model names (claude-3-5-sonnet-20241022, minicpm-v:8b, gpt-4o-mini-2024-07-18, pixtral-large-2411, pixtral-12b-2409, llama3.2-vision:11b, llava:7b)
- **Y-axis:** Overall score (average across all criteria)

Bar chart showing the overall performance of each model. It provides a clear ranking of the models based on their ability to capture and describe the visual aspects.

The evaluation process yielded a comprehensive set of scores for each model across the seven criteria. These scores, presented in Table 1, provide a detailed overview of the models' performance in capturing different visual aspects:

**Table 1: Model Performance**

| Model | Main Object | Additional Objects | Background | Detail | Dominant Colors | Style | Viewpoint |
|---|---|---|---|---|---|---|---|
| 1 | 67.23 | 66.03 | 59.80 | 74.76 | 67.10 | 74.18 | 75.73 |
| 2 | 60.98 | 62.17 | 55.30 | 75.00 | 62.91 | 74.21 | 72.77 |
| 3 | 63.02 | 63.24 | 55.03 | 72.22 | 57.43 | 80.32 | 68.13 |
| 4 | 54.67 | 63.06 | 49.39 | 73.00 | 59.36 | 74.31 | 67.86 |
| 5 | 54.81 | 60.19 | 48.22 | 72.11 | 55.69 | 72.39 | 66.81 |
| 6 | 47.50 | 56.39 | 31.61 | 71.58 | 45.63 | 66.22 | 50.50 |
| 7 | 41.99 | 32.82 | 33.59 | 71.30 | 43.08 | 68.18 | 45.75 |

Model Key:

1. claude-3-5-sonnet-20241022
2. minicpm-v:8b
3. gpt-4o-mini-2024-07-18
4. pixtral-large-2411
5. pixtral-12b-2409
6. llama3.2-vision:11b
7. llava:7b

To facilitate a clearer comparison, the overall performance of each model, calculated by averaging the scores across all criteria, is presented in Table 2:

**Table 2: Overall Model Performance**

| Model | Overall Score |
|---|---|
| claude-3-5-sonnet-20241022 | 69.26 |
| gpt-4o-mini-2024-07-18 | 65.63 |
| llama3.2-vision:11b | 52.78 |
| llava:7b | 48.10 |
| minicpm-v:8b | 66.19 |
| pixtral-12b-2409 | 61.46 |
| pixtral-large-2411 | 63.09 |

## 5. KEY OBSERVATIONS

- claude-3-5 consistently performs best or near-best across most criteria, especially strong in Background, Dominant Colors, Main Object, Viewpoint, and Additional Objects.
- gpt-4o-mini stands out for Style with the highest individual score (80.32).
- minicpm-v:8b excels at Detail, topping out at 75.00.

The criteria with the highest overall success rates (Style, Detail, and Viewpoint) may indicate tasks where these models have stronger predictive or interpretive capabilities.

The relatively lower scores for Background and Dominant Colors suggest these aspects are more challenging for all models or possibly less consistently evaluated.

These patterns can help guide which model to pick for a specific type of task (for example, using gpt-4o-mini if "Style" is crucial, or minicpm-v for intricate "Detail").

## 6 SUMMARY AND CONCLUSIONS

This research presented a novel benchmark for evaluating the performance of multimodal models in fine-grained image analysis. By systematically combining diverse visual aspects, a dataset of 14,580 images was generated to assess the models' ability to capture and describe specific visual features. Seven leading multimodal models were evaluated across seven criteria: main object, additional objects, background, detail, dominant colors, style, and viewpoint.

The results revealed significant variations in performance across models and criteria. Claude-3-5-sonnet-20241022 emerged as the top-performing model overall, demonstrating consistent accuracy across most criteria. Notably, gpt-4o-mini-2024-07-18 excelled in capturing style, while minicpm-v:8b showed superior performance in identifying details. The analysis also highlighted areas where models generally struggled, such as accurately interpreting backgrounds and dominant colors.

This benchmark provides valuable insights for researchers and developers working with multimodal models. By identifying the strengths and weaknesses of different models, it can guide the selection of appropriate models for specific tasks and inform future research directions. The findings emphasize the need for continued development in areas such as background and color analysis to further enhance the capabilities of multimodal models for comprehensive image understanding.

This research contributes to a deeper understanding of the capabilities and limitations of current multimodal models in fine-grained image analysis. The benchmark and its findings serve as a valuable resource for the research community, paving the way for the development of more robust and versatile multimodal models for various image understanding applications.

**AUTHOR BIOGRAPHIES**

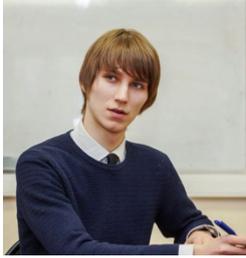Evgenii Evstafev is a skilled software developer at the University of Cambridge, where he has been working since September 2022, specializing in identity and access management. He earned a Bachelor's degree in Business Informatics from the Higher School of Economics (2010-2014) and a Master's degree in Programming from Perm National Research Polytechnic University (2014-2016). Evgenii also taught at the Faculty of Mechanics and Mathematics at Perm State University while engaged in postgraduate studies (2016-2019). His professional journey spans over 11 years across various industries, including roles such as System Architect at L'Etoile (2021-2022) focusing on product development, the Head of Analytics at Gazprombank (2020-2021), and Head of the Department for System Analysis and Software Design at Center 2M (2019-2020). Additionally, he worked on system development at the energy company T Plus (2016-2019).